\documentclass{article}

\usepackage[preprint]{neurips_2025}

\usepackage[utf8]{inputenc} 
\usepackage[T1]{fontenc}    
\usepackage{hyperref}       
\usepackage{url}            
\usepackage{booktabs}       
\usepackage{amsfonts}       
\usepackage{nicefrac}       
\usepackage{microtype}      
\usepackage{xcolor}         
\usepackage{amsmath}
\usepackage{tcolorbox}
\usepackage{cleveref}
\usepackage{setspace}
\usepackage{xspace}

\usepackage{algorithm}
\usepackage{algorithmic}
\usepackage{caption}
\title{Reinforcement Learning for Flow-Matching Policies}

\author{%
  Samuel Pfrommer, Yixiao Huang, Somayeh Sojoudi \\
  Department of Electrical Engineering and Computer Sciences \\
  University of California, Berkeley \\
  Berkeley, CA 94720 \\
  \texttt{\{sam.pfrommer,yixiaoh,somayeh\}@berkeley.edu} \\
  \thanks{
    Code available at \url{https://github.com/spfrommer/flowmatching_policy_rl}.
  }
}

\newcommand{\bg}{\mathbf{g}}
\newcommand{\bp}{\mathbf{p}}
\newcommand{\bh}{\mathbf{h}}
\newcommand{\bu}{\mathbf{u}}
\newcommand{\bv}{\mathbf{v}}

\newcommand{\ilfm}{\textcolor[HTML]{120789}{\textup{\textbf{ILFM}}}\xspace}
\newcommand{\rwfm}{\textcolor[HTML]{C23D80}{\textup{\textbf{RWFM}}}\xspace}
\newcommand{\grpo}{\textcolor[HTML]{FA9E3B}{\textup{\textbf{GRPO}}}\xspace}

\newcommand{\R}{\mathbb{R}}
\newcommand{\E}{\mathbb{E}}
\newcommand{\Unif}{\mathcal{U}}
\newcommand{\Normal}{\mathcal{N}}
\newcommand{\ones}{\mathbf{1}}

\newcommand{\rollout}{\text{Rollout}}

\newcommand{\state}{\mathbf{s}}

\newcommand{\aobs}{\tilde{\mathbf{o}}}
\newcommand{\obs}{\mathbf{o}}
\newcommand{\Obs}{\mathbf{O}}

\newcommand{\act}[1][t]{\mathbf{a}_{#1}}
\newcommand{\Act}{\mathbf{A}}

\newcommand{\data}{\mathcal{D}}
\newcommand{\datademo}{\mathcal{D}_{D}}
\newcommand{\datapolicy}{\mathcal{D}_{\pi}}

\newcommand{\supp}{\text{Supp}}

\newcommand{\inlinesubsection}[1]{\textbf{#1.}}

\DeclareMathOperator*{\argmax}{arg\,max}

\newtheorem{definition}{Definition}

\begin{document}

\maketitle

\begin{abstract}
Flow-matching policies have emerged as a powerful paradigm for generalist robotics.
These models are trained to imitate an action chunk, conditioned on sensor observations and textual instructions.
Often, training demonstrations are generated by a suboptimal policy, such as a human operator.
This work explores training flow-matching policies via reinforcement learning to surpass the original demonstration policy performance.
We particularly note minimum-time control as a key application and present a simple scheme for variable-horizon flow-matching planning.
We then introduce two families of approaches: a simple Reward-Weighted Flow Matching (\rwfm) scheme and a Group Relative Policy Optimization (\grpo) approach with a learned reward surrogate.
Our policies are trained on an illustrative suite of simulated unicycle dynamics tasks, and we show that both approaches dramatically improve upon the suboptimal demonstrator performance, with the \grpo approach in particular generally incurring between $50\%$ and $85\%$ less cost than a naive Imitation Learning Flow Matching (\ilfm) approach.
\end{abstract}

\section{Introduction}

Flow-matching policies are commonly employed to build Vision-Language-Action (VLA) models, which have shown great promise in providing flexible visuomotor behavior \citep{brohan2023rt, kim2024openvla, driess2023palme, black2024pi0}. Where traditional robotic machine learning solutions are only exposed to task-specific data, VLAs are trained on a large multi-embodiment corpus of human demonstrations using Imitation Learning (IL) \citep{brohan2023rt,black2024pi0}.
Modern VLAs typically accomplish this by using a diffusion or flow matching action expert to generate a fixed-duration action chunk, conditioned on sensor observations and textual instructions \citep{black2024pi0,octo2024octo, shukor2025smolvla}.
This architecture allows for continuous control in high-dimensional action spaces across a diverse set of environments.

However, the performance of current VLAs is limited by demonstration data quality.
Human operators are often inconsistent in their demonstrations, leading to unwanted variability in the imitating policy \citep{xu2022discriminator}.
Furthermore, even the best demonstration trajectories are often suboptimal, especially with regard to task completion speed.
This work aims to improve policies beyond the performance of the demonstration data.

Formulating this generally requires access to a reward function which applies to a range of embodiments and tasks. 
Such a reward function might consist of task completion scores from a VLM evaluation model \citep{li2025self,ma2024vision} as well as general robotics desiderata such as minimum-time control.
Our work does not assume a particular reward function; we instead evaluate our reinforcement learning method with a range of hand-designed rewards, including time-minimizing and actuation-minimizing components as well as other illustrative objectives.

Minimum-time reinforcement learning is incompatible with existing VLAs, which inherit diffusion-based planners' weakness of only being able to generate fixed-horizon action chunks \citep{janner2022planning,chi2023diffusionpolicy}.
This fixed planning horizon frequency leads to inefficient trajectories, such as unnecessary back-and-forth movements to consume superfluous planning time \citep{lee2023refining,chen2024simple,luo2025generative}.
Our work addresses this with a simple incorporation of planning horizon into the flow matching action chunk.
This allows us to introduce a novel online RL framework for flow matching policies that enables optimization for minimum-time control as well as a range of other illustrative RL objectives.

\subsection{Related Work}

\inlinesubsection{RL for diffusion models} 
While RL has been increasingly used to fine-tune diffusion models with user-defined rewards \citep{fan2023dpok,black2023training,wallace2024diffusion,ren2024diffusion}, these techniques are challenging to extend to continuous flow-based models. Most existing methods rely on the variational Evidence Lower Bound (ELBO) \citep{ho2020denoising} to approximate the policy likelihood, which then serves as a reward signal. However, applying this approach to flow-based models is computationally prohibitive due to the need for expensive divergence-trace estimation \citep{lipman2022flow}. Although some recent methods bypass explicit likelihood computation, they remain incompatible with our setup---some require differentiable reward functions \citep{domingo2024adjoint}, while others depend on filtered offline data \citep{dong2023raft} which is not desired due to the online-offline gap \citep{tang2024understanding}. 

\inlinesubsection{RL for flow matching in other domains} Recent studies have applied RL to align flow-based models with human preferences across diverse domains such as image synthesis \citep{fu2025chats}, video generation \citep{liu2025improving}, and protein design \citep{huguet2024sequence}.  Notably, both \citep{huguet2024sequence,liu2025improving} incorporate a reward-weighted regression (RWR) objective with flow-based models.
To overcome the offline data gap while avoiding the high computation cost of likelihood estimation, \citet{fan2025online} proposes an online reward-weighted fine-tuning method for flow-based models. Recently, \citet{shao2024deepseekmath,guo2025deepseek} introduce Group Relative Policy Optimization (GRPO), a Proximal Policy Optimization (PPO) \citep{schulman2017proximal} variant that eliminates the value model and thus reduces memory usage and computational overhead. Building on this idea, \citet{xue2025dancegrpo} extends GRPO to flow-based models, demonstrating its promise for visual generation tasks.

\inlinesubsection{Online learning for VLAs} Recent work has explored online RL for VLAs \citep{julg2025refined,guo2025improving,ramrakhya2025grounding}.
\citet{guo2025improving} apply an on-policy actor-critic method to fine-tune the action head of a VLA model. However, their work focuses on single-step action prediction and is not applicable to modern diffusion-based models.
\citet{julg2025refined} distill a generalist VLA policy into a student model via PPO \citep{schulman2017proximal}. Similarly, \citet{ramrakhya2025grounding} propose to use LLM-generated rewards to fine-tune the VLA policy with PPO.
None of these works are directly applicable to improving a modern diffusion-based VLA model with RL, which is the focus of this work.

\inlinesubsection{Fixed-horizon planning} 
Diffusion-based models for robot planning \citep{janner2022planning,chi2023diffusionpolicy} require an a priori selection of the planning horizon.
This generally leads to infeasible or unsuccessful trajectories in the case of an underestimated horizon and inefficient trajectories in the case of an overestimated horizon \citep{janner2022planning,lee2023refining,chen2024simple,luo2025generative}.
Even hierarchical and stitching-based approaches inherit this issue, where a priori selection of the planning horizon is replaced with a priori selection of the number of stitched trajectories \citep{luo2025generative, hao2025chd}.
Existing approaches which claim to enable variable-horizon planning suffer serious drawbacks.
\citet{he2024rediffuser} proposes planning over many candidate horizons and selecting the appropriate horizon post hoc with a confidence estimator, a computationally expensive and complicated process.
\citet{kim2024stitching} introduces a stitching-based approach to generates goal-seeking subtrajectories, guided by an external value function. This does not truly enable variable-horizon planning, as the horizon is constrained to be a multiple of the subtrajectory length and must be iteratively generated along with execution.
Approaches which alternate between high-level goal prediction and low-level diffusion planning share this weakness \citep{ma2024hierarchical}.
\citet{chen2024diffusion} combine teacher forcing and diffusion denoising to produce dynamically feasible trajectories over a range of horizons, but the desired horizon still must be selected a priori.

\subsection{Contributions}
We make the following contributions:
\begin{enumerate}
    \item We formalize the flow matching policy RL problem and contextualize the importance of overcoming variation and support suboptimality when learning from demonstrations.
    \item We incorporate action chunk time horizons in the flow matching problem, allowing for adaptive control of the trajectory horizon when generating.
    \item We introduce a Reward-Weighted Flow Matching (\rwfm) scheme with a novel action trajectory explorer that addresses both variation and support suboptimality.
    \item We improve sample efficiency using a Group Relative Policy Optimization (\grpo) approach with a learned reward surrogate.
\end{enumerate}

\subsection{Notation}

We let $\R$ denote the real numbers and write vectors $\mathbf{x} \in \R^n$ and vector-valued functions using boldface. We write a $n$-dimensional vector of all ones as $\ones_n \in \R^n$ and the identity matrix as $\mathbf{I}_n \in \R^{n \times n}$. Expectation is denoted as $\E$ and the $n$-dimensional standard multivariate normal distribution as $\Normal(\mathbf{0}_n, \mathbf{I}_n)$. The uniform distribution on an interval is written as $\Unif(a, b)$.
\section{Problem}

We consider a partially observable environment featuring an underlying system state $\state$ and observation $\obs$.
We aim to learn a policy $\pi_{\theta}(\Act \mid \aobs)$, where $\Act = [\act[1], \act[2], \ldots, \act[H]]$ is an action chunk and the augmented observation $\aobs$ combines $\obs$ with an external command (e.g., a target position vector or a language instruction).
Given a state $\state$ and action chunk $\Act$, we define a rollout function $\rollout(\state, \Act)$ that returns a trajectory of observations $\Obs$, where $\Obs = [ \obs_1, \obs_2, \ldots, \obs_{H+1} ]$.

As we restrict our attention to the visuomotor problem, we assume a distribution $(\state, \aobs) \sim p_{\state, \aobs}$ over state-observation pairs and do not consider sequential decision-making.
We also assume access to a scalar-valued reward function $R(\aobs, \Act, \Obs)$.
While we consider hand-specified reward functions in this work, we note that we do not impose any constraints on the form of $R$.
In future work, it could be possible to learn $R$ or derive it from a VLM evaluation model \citep{li2025self,ma2024vision}.

\inlinesubsection{Expert suboptimality}
We consider a demonstration policy $\pi_D(\Act \mid \aobs)$ whose actions are suboptimal due to both variation and limited support. Formally, assume $\pi_D(\,\cdot \mid \aobs)$ admits a density function $p_D(\Act \mid \aobs)$ and let $\supp_{\aobs} = \{ \Act \mid p_D(\Act \mid \aobs) > 0 \}$. 

\emph{Variation suboptimality} can be captured as an inequality condition
\begin{align}
   \argmax_{\Act \in \supp_{\aobs}} R\big(\aobs, \Act, \rollout(\state, \Act)\big) -
   \E_{\Act \sim \pi_{\theta}(\Act \mid \aobs)} 
   \left[ R\big(\aobs, \Act, \rollout(\state, \Act)\big) \right]
   > \delta
\end{align}
for any particular state-observation pair $(\state, \aobs)$ and some $\delta > 0$.

\emph{Support suboptimality} is captured via
\begin{align}
   \argmax_{\Act} R\big(\aobs, \Act, \rollout(\state, \Act)\big) - \argmax_{\Act \in \supp_{\aobs}} R\big(\aobs, \Act, \rollout(\state, \Act)\big) > \delta
\end{align}
for any particular state-observation pair $(\state, \aobs)$ and some $\delta > 0$.

We emphasize that both of these distinct forms of suboptimality occur frequently in demonstration datasets. A human demonstrator will exhibit natural variance in their ability to complete a task, with some attempts achieving better reward than others. Furthermore, there are generally superior control strategies which exceed even the best efforts of a human demonstrator.

\inlinesubsection{Imitation learning pretraining}
We first introduce a standard imitation learning approach for flow matching policies, disregarding the reward function \citep{black2024pi0}.
This work assumes access to a paired dataset $\datademo = \{(\aobs^{(i)}, \Act^{(i)}, \Obs^{(i)}) \}_{i=1}^{N}$ with augmented observations $\aobs$ sampled from the marginal of $p_{\state, \aobs}$ and actions sampled from a demonstration policy $\pi_D$. Observation trajectories $\Obs$ are produced via rollouts of $\Act$. We do not assume access to the underlying state $\state$, which may not be available in practice. Our training objective is then to minimize the following conditional flow matching loss:

\begin{definition}[\ilfm loss]
    \label{def:ilfm_loss}
    For a trajectory dataset $\data = \{(\aobs^{(i)}, \Act^{(i)}, \Obs^{(i)}) \}_{i=1}^{N}$, we define the Imitation Learning Flow Matching (\ilfm) loss as
\begin{align} \label{eq:ilfm_loss}
    \mathcal{L}(\theta) = \E_{
        (\aobs, \Act, \Obs) \sim \datademo,
        \tau \sim \Unif(0, 1),
        \Act^{\tau} \sim p^{\tau}( \cdot \mid \Act)
    }
    \left[ \| \bv_{\theta}(\Act^{\tau}, \aobs, \tau) - \bu(\Act^{\tau} \mid \Act) \|^2 \right],
\end{align}
where $\tau$ denotes the flow-matching time step, $p^{\tau}(\,\cdot \mid \Act) \sim \Normal(\tau \Act, (1 - \tau) \mathbf{I})$ is the optimal transport probability path, $\bu$ is the denoising vector field, and $\bv_{\theta}$ is the learned vector field.
\end{definition}
We evaluate \eqref{eq:ilfm_loss} in practice by sampling $\epsilon \sim \Normal(\mathbf{0}, \mathbf{I})$, letting $\Act^{\tau} = \tau \Act + (1 - \tau) \epsilon$, and regressing onto $\bu(\Act^{\tau} \mid \Act) = \epsilon - \Act$.
To generate using $\pi_{\theta}(\Act \mid \aobs)$, we sample random noise $\Act^0 \sim \Normal(\mathbf{0}, \mathbf{I})$, and integrate along the learned vector field $\bv_{\theta}$ from $\tau=0$ to $\tau=1$ to obtain $\Act^1$.

\inlinesubsection{Reward-guided posttraining}
Learning $\pi_{\theta}$ purely via imitation will reproduce suboptimal demonstrations. 
To address this, we aim to instead maximize the expected reward of the action trajectories sampled from our policy:
\begin{align} \label{eq:reward_loss}
    \argmax_{\theta} \; \E_{
        \state, \aobs \sim p_{\state, \aobs},
        \Act \sim \pi_{\theta}(\Act \mid \aobs)
    } \left[ R\big(\aobs, \Act, \rollout(\state, \Act)\big) \right].
\end{align}
Note that in theory the probability density of $\Act$ under $\pi_{\theta}$ can be evaluated by integrating $\bv_{\theta}$ from $\tau=1$ to $\tau=0$, accumulating divergence terms. However, this requires expensive simulation and the evaluation of a vector-Jacobian product at each numerical integration step. We now introduce an alternative approach to optimizing \eqref{eq:reward_loss}.

\section{Method} \label{sec:method}

We explore two distinct families of RL approaches.
The first applies recent results from Reward-Weighted Flow Matching (\rwfm) to selectively imitate high-reward demonstrations. While this alone is effective for addressing variation suboptimality, we argue that it is not sufficient for addressing support suboptimality and correspondingly introduce an action explorer into the \rwfm framework.
Our second approach combines Group Relative Policy Optimization (\grpo) with a learned reward surrogate, which we show to be more sample efficient than \rwfm.

For both approaches, we are interested in generating trajectories of variable duration. Time is at a premium in many real-world applications, and reinforcement learning is required to exceed the speed of suboptimal human demonstrators. Our training procedure discussions are thus preceded by a simple framework for generating variable-duration trajectories with flow matching models.

Our training setup shares a common structure between the \rwfm and \grpo methods. We are primarily interested in evaluating \emph{sample efficiency}, or how many collected trajectories are required to achieve a particular level of performance. At a high level, we train each approach on the demonstration dataset until validation performance stagnates. We then collect a new batch of trajectories of size proportional to the number of collected trajectories so far and continue training.

\inlinesubsection{Variable-time planning with diffusion and flow matching models}
Conventional approaches for planning with diffusion and flow matching models involve discretizing a fixed time duration. For example, the $\pi_0$ model \citep{black2024pi0} uses a time duration of $2$ seconds discretized into a horizon $H=50$. This results in conservative trajectories for easy objectives and prematurely truncated trajectories for long-horizon objectives. We now enable a flow matching model to generate variable-time samples with minimal changes to the standard velocity-field U-Net architecture.

Consider a dataset of demonstrations $\data = \{(\aobs^{(i)}, \Act^{(i)}, \Obs^{(i)}) \}_{i=1}^{N}$ where each action chunk $\Act^{(i)} \in \R^{d_a \times H_i}$ has a different horizon $H_i$, all with a constant time step $\Delta t$ determined by the expert control frequency. While U-Nets can process inputs of variable length, the output horizon is always equivalent to the input horizon, requiring a priori selection of the trajectory horizon $H$ and thus fixing the total time duration as $H \Delta t$ \citep{black2024pi0}.

We address this by 1) interpolating action chunks to a fixed horizon $H'$ and 2) concatenating an additional channel which encodes the original horizon $H$.
For concreteness, consider an action tensor $\Act \in \R^{d_a \times H}$, where each $\act[i] \in \R^{d_a}$.
An evenly-spaced linear interpolation to a fixed horizon $H'$ produces $\Act' \in \R^{d_a \times H'}$.
We then extend $\Act'$ to $\hat{\Act} \in \R^{d_a+1 \times H'}$ by concatenating $\Act$ with $H \cdot \ones_{d_a}$.
The resulting augmented action chunk $\hat{\Act}$ can be considered as a standard denoising target when training a U-Net, after applying appropriate normalization to each channel.

When generating, we sample random noise in $\R^{d_a+1 \times H'}$ and integrate along the U-Net velocity field to produce an augmented action chunk $\hat{\Act} \in \R^{d_a+1 \times H'}$. This must now be interpolated back to the original control frequency of the expert demonstrations. We extract the desired horizon $H$ by taking the mean of channel $d_a+1$ of $\hat{\Act}$ and rounding to the nearest integer. This channel is then discarded, and the remaining tensor is linearly interpolated to a chunk of horizon $H$. We have thus enabled our flow matching model to generate variable-duration trajectories.

For the remainder of this work, references to $\Act$ generally refer to the time-augmented action chunk $\hat{\Act} \in \R^{d_a + 1 \times H'}$ for notational simplicity.

\subsection{Reward-weighted flow matching} \label{sec:rwfm}
We consider a demonstration policy $\pi_D(\Act \mid \obs)$ whose actions are both variation-suboptimal and support-suboptimal. This section first addresses variation suboptimality by introducing a reward-weighted flow matching approach to subselect high-performing expert demonstrations. We then extend action sampling beyond the support of the demonstration by introducing a simple action exploration distribution.

\inlinesubsection{Training procedure}
We introduce recent key results from the flow matching literature into our framework. The following result intuitively states that weighted flow matching produces a learned distribution with density proportional to the weighting function.

\begin{tcolorbox}[title=Weighted Flow Matching (Informal -- \citep{fan2025online})]
    Let $q(x)$ be the data distribution and $w(x)$ be a non-negative weighting function. Then under the weighted flow matching objective
    \begin{align} \label{eq:weighted_flow_matching_loss}
        \mathcal{L}_{w}(\theta) = \E_{
            x \sim q(x),
            \tau \sim \Unif(0, 1),
            x^{\tau} \sim p^{\tau}( \cdot \mid x)
        }
        \left[ w(x)\,\| \bv_{\theta}(x^{\tau}, \tau) - \bu(x^{\tau} \mid x) \|^2 \right],
    \end{align}
    the distribution $p_{\theta^*}$ obtained by training $\bv_{\theta}$ to optimality under \eqref{eq:weighted_flow_matching_loss} has the density
    \begin{align} \label{eq:weighted_flow_matching_density}
        p_{\theta^*}(x) \propto w(x) q(x).
    \end{align}
    Further consider an iterative iterative training approach, where instead of sampling $x \sim q(x)$ in \eqref{eq:weighted_flow_matching_loss}, we sample from the learned model $p_{\theta^*}^E$ after $E$ epochs of training to generate training data for $p_{\theta^*}^{E+1}$. Then we arrive at the density
    \begin{align} \label{eq:weighted_flow_matching_density_iterative}
        p_{\theta^*}^E(x) \propto \frac{w(x)^E q(x)}{Z_E},
    \end{align}
    where $Z_E$ is a normalizing constant.
\end{tcolorbox}

Note that repeated iterative sampling leads the density \eqref{eq:weighted_flow_matching_density_iterative} to converge to a Dirac distribution centered at $\argmax_{x} w(x)$.
This poses a diversity problem in generative image models, which \citet{fan2025online} address by introducing Wasserstein regularization. However, for our application, it is in fact desirable for the policy to converge to a reward-maximizing trajectory conditional on the observation.
This result suggests a basic strategy for re-weighting expert demonstrations to emphasize high-reward action trajectories.

\begin{definition}[\rwfm loss]
    \label{def:rwfm_loss}
    For a trajectory dataset $\data = \{(\aobs^{(i)}, \Act^{(i)}, \Obs^{(i)}) \}_{i=1}^{N}$, we define the Reward-Weighted Flow Matching (\rwfm) loss as
\begin{align} \label{eq:rwf_loss}
    \mathcal{L}(\theta) = \E_{
        (\aobs, \Act, \Obs) \sim \data,
        \tau \sim \Unif(0, 1),
        \Act^{\tau} \sim p^{\tau}( \cdot \mid \Act)
    }
    \left[ w(\aobs, \Act, \Obs) \| \bv_{\theta}(\Act^{\tau}, \aobs, \tau) - \bu(\Act^{\tau} \mid \Act) \|^2 \right],
\end{align}
    where $w(\aobs, \Act, \Obs) = \exp(\alpha R(\aobs, \Act, \Obs)) > 0$ for a scaling parameter $\alpha > 0$. The remaining notation is as in \Cref{def:ilfm_loss}.
\end{definition}

Note that for $\alpha=0$, the \rwfm loss reduces to the \ilfm loss.

During training, we alternate between expanding the training dataset by collecting samples from the current policy and minimizing \eqref{eq:rwf_loss}. We collect more samples when the validation reward plateaus for a certain number of epochs. Our algorithm is summarized in \Cref{alg:rwfm}. The $\text{plateau}$ function depends on $r_{val}$, $r_{best}$, the current epoch, and the epoch where the best validation performance was achieved. We omit the latter two arguments for brevity. In practice, we evaluate the expectation for $r_{val}$ empirically by sampling $(\state, \aobs)$ pairs from a held out constant validation set.

\begin{algorithm}
    \caption{\rwfm training loop}
    \label{alg:rwfm}
\setstretch{1.15}
\begin{algorithmic}
\vspace{0.5\baselineskip}
\INPUT{Collection iterations $C$, collection sampling fraction $\gamma$, expert training demonstrations $\datademo$}
\OUTPUT{policy $\pi_{\theta}$}
\vspace{0.5\baselineskip}
\STATE \textbf{initialize} flow-matching model $\bv_{\theta}(\Act^{\tau}, \aobs, \tau)$

\vspace{0.5\baselineskip}

\STATE $\data_{train} \gets \datademo$
\vspace{0.5\baselineskip}
\FOR{\_ $\gets$ $1$ \textit{to} $C$}
    \STATE $r_{best} \gets -\infty$
    \STATE $r_{val} \gets
    \E_{(\state, \aobs) \sim p_{\state, \aobs}, \Act \sim \pi_{\theta}(\Act \mid \aobs)}
    \left[ R\big(\aobs, \Act, \rollout(\state, \Act)\big) \right]$

    \vspace{0.5\baselineskip}
    \WHILE{\emph{not} plateau($r_{val}$, $r_{best}$)}
        \STATE \textbf{train} $\bv_{\theta}$ for one epoch of \rwfm loss (\Cref{def:rwfm_loss}) over $\data_{train}$
        \STATE $r_{val} \gets
        \E_{(\state, \aobs) \sim p_{\state, \aobs}, \Act \sim \pi_{\theta}(\Act \mid \aobs)}
        \left[ R\big(\aobs, \Act, \rollout(\state, \Act)\big) \right]$
        
        \STATE $r_{best} \gets \max(r_{val}, r_{best})$
    \ENDWHILE

    \vspace{0.5\baselineskip}
    \STATE $\datapolicy \gets \left\{ (\aobs^{(i)}, (\Act')^{(i)}, \Obs^{(i)})\right\}_{i=1}^{\gamma |\data_{train}|}$ with \\[0.05cm]
    \qquad $\state, \aobs \sim p_{\state, \aobs}$,
    $\Act \sim \pi_{\theta}(\Act \mid \aobs)$,
    $\Act' \sim \mathcal{E}\big( \Act' \mid \Act\big)$,
    $\Obs \gets \rollout\big(\state, \Act'\big)$
    \vspace{0.3\baselineskip}

    \STATE $\data_{train} \gets \data_{train} \cup \datapolicy$
\ENDFOR
\end{algorithmic}
\end{algorithm}

\inlinesubsection{Action exploration}
\Cref{eq:weighted_flow_matching_density_iterative} reveals a key drawback: if the data distribution $q(x)$ has zero density at some point $x$, that point will maintain a null density regardless of the reward weight $w(x)$. This is natural for image generation, where the goal is to focus generation on a subregion of the data manifold. The implications for trajectory generation are less desirable. Specifically, support suboptimality suggests that a potentially superior action trajectory might never be explored if it lies outside the abilities of the expert demonstrator.

Traditional continuous-action online RL methods encourage exploration by sampling actions from a normal distribution when collecting episodes. We aim to extend this to our setting, which involves action trajectories. Simply adding i.i.d. noise to the action trajectories produces non-smooth trajectories and does not efficiently explore the action space.

We instead introduce a simple action exploration distribution $\mathcal{E}(\Act' \mid \Act)$ that adds random ``bumps'' to each channel of the action trajectory independently. Specifically, bump centers are uniformly sampled across the trajectory horizon, widths range from $H/16$ to $H/4$, and amplitudes are randomly sampled from $\Unif(-M, M)$ for a magnitude parameter $M > 0$. These Gaussian-shaped perturbations are added to the original trajectory and the resulting sum is clamped to maintain valid action bounds. This perturbation process is applied equivalently to the action and time channels of the augmented action chunk $\hat{\Act}$.

For our \rwfm implementation, we use the action explorer when collecting rollouts. This differs from the \grpo approach which we now introduce.

\subsection{Group relative policy optimization} \label{sec:grpo}

\begin{algorithm}
    \caption{\grpo training loop}
    \label{alg:grpo}
\setstretch{1.15}
\begin{algorithmic}
\vspace{0.5\baselineskip}
\INPUT{Pretraining epochs $E_{pt}$, reward surrogate training sub-epochs $E_{rs}$, collection iterations $C$, collection sampling fraction $\gamma$, expert demonstrations $\datademo$}
\OUTPUT{policy $\pi_{\theta}$}
\vspace{0.5\baselineskip}
\STATE \textbf{initialize} flow-matching model $\bv_{\theta}(\Act^{\tau}, \aobs, \tau)$
\STATE \textbf{initialize} reward surrogate $R_{\phi}(\aobs, \Act)$
\FOR{\_ $\gets$ $1$ \textit{to} $E_{pt}$}
    \STATE \textbf{train} $\bv_{\theta}$ for one epoch of \ilfm loss (\Cref{def:ilfm_loss}) over $\data_{D}$
    \STATE \textbf{train} $R_{\phi}$ for $E_{rs}$ epochs of reward surrogate loss (\Cref{def:reward_surrogate_loss}) over $\data_{D}$
\ENDFOR
\vspace{0.5\baselineskip}

\STATE $\data_{train} \gets \datademo$
\vspace{0.5\baselineskip}
\FOR{\_ $\gets$ $1$ \textit{to} $C$}
    \STATE $r_{best} \gets -\infty$
    \STATE $r_{val} \gets
    \E_{(\state, \aobs) \sim p_{\state, \aobs}, \Act \sim \pi_{\theta}(\Act \mid \aobs)}
    \left[ R\big(\aobs, \Act, \rollout(\state, \Act)\big) \right]$

    \vspace{0.5\baselineskip}
    \WHILE{\emph{not} plateau($r_{val}$, $r_{best}$)}
        \STATE \textbf{train} $\bv_{\theta}$ for one epoch of \grpo loss (\Cref{def:grpo_loss}) using $\mathcal{E}(\Act' \mid \Act)$ over $\data_{train}$
        \STATE \textbf{train} $R_{\phi}$ for $E_{rs}$ epochs of reward surrogate loss (\Cref{def:reward_surrogate_loss}) over $\data_{train}$
        \STATE $r_{val} \gets
        \E_{(\state, \aobs) \sim p_{\state, \aobs}, \Act \sim \pi_{\theta}(\Act \mid \aobs)}
        \left[ R\big(\aobs, \Act, \rollout(\state, \Act)\big) \right]$
        
        \STATE $r_{best} \gets \max(r_{val}, r_{best})$
    \ENDWHILE

    \vspace{0.5\baselineskip}
    \STATE $\datapolicy \gets \left\{ (\aobs^{(i)}, (\Act')^{(i)}, \Obs^{(i)})\right\}_{i=1}^{\gamma |\data_{train}|}$ with \\[0.05cm]
    \qquad $\state, \aobs \sim p_{\state, \aobs}$,
    $\Act \sim \pi_{\theta}(\Act \mid \aobs)$,
    $\Obs \gets \rollout\big(\state, \Act\big)$
    \vspace{0.3\baselineskip}

    \STATE $\data_{train} \gets \data_{train} \cup \datapolicy$
\ENDFOR
\end{algorithmic}

\end{algorithm}

This section improves the sample efficiency of the policy learning process described in \Cref{sec:rwfm} by introducing a GRPO-based method with a learned reward surrogate $R_{\phi}(\aobs, \Act)$, which critically does not rely on collecting a rollout to obtain an observation trajectory. Our high-level approach involves sampling $G$ action chunks from the current policy and evaluating $R_{\phi}$ to produce a set of advantages. We then use the advantages in the weighted flow matching loss. We incorporate the action explorer $\mathcal{E}(\Act' \mid \Act)$ into the policy sampling step to overcome support suboptimality as in \Cref{sec:rwfm}. Thus, unlike \rwfm, we do not employ $\mathcal{E}$ when collecting rollouts for \grpo. This allows us to validate diverse explored actions against a reward model without running potentially expensive rollouts.

\begin{definition}[\grpo loss]
    \label{def:grpo_loss}
    For an augmented observation dataset $\data = \{\aobs^{(i)} \}_{i=1}^{N}$ and reward surrogate $R_{\phi}(\aobs, \Act)$, we define the Group Relative Policy Optimization (\grpo) loss as
\begin{align} \label{eq:grpo_loss}
    \begin{split}
    \mathcal{L}(\theta) = &\E_{
        \aobs \sim \data,
        \Act_i \dots \Act_G \sim \pi_{\theta}(\Act \mid \aobs),
        \Act_i' \sim \mathcal{E}(\Act' \mid \Act_i),
        \tau \sim \Unif(0, 1),
        (\Act_i')^{\tau} \sim p^{\tau}( \cdot \mid \Act_i')
    } \\
    &\quad\frac{1}{G}
    \sum_{i=1}^{G}
    \left[ w(\aobs, \Act_i') \left\| \bv_{\theta}\big((\Act_i')^{\tau}, \aobs, \tau\big) - \bu\big((\Act_i')^{\tau} \mid \Act_i'\big) \right\|^2 \right],
    \end{split}
\end{align}
    where $w(\aobs, \Act_i') = \exp(\alpha a_i) > 0$ for a scaling parameter $\alpha > 0$ and advantages $a_i$. Advantages are computed from rewards $r_i = R_{\phi}(\aobs, \Act_i')$ via
\[
    a_i = \frac{r_i - \text{mean}(\{r_1, \dots, r_G\})}{\text{std}(\{r_1, \dots, r_G\}) + \epsilon}.
\]
The remaining notation is as in \Cref{def:ilfm_loss}.
\end{definition}

If training is performed in simulation, the reward surrogate could be replaced with a rollout followed by an application of the reward function: $R\big(\aobs, \Act, \rollout(\state, \Act)\big)$. This requires resetting the simulator to a particular state $G$ times and performing a separate rollout for each action chunk. As this is unrealizable in the real world, we instead use a learned reward surrogate $R_{\phi}(\aobs, \Act)$ which is regressed on previously observed rewards in the training dataset.

\begin{definition}[Reward surrogate loss]
    \label{def:reward_surrogate_loss}
    For a trajectory dataset $\data = \{(\aobs^{(i)}, \Act^{(i)}, \Obs^{(i)}) \}_{i=1}^{N}$, we define the reward surrogate loss as
\begin{align} \label{eq:rs_loss}
    \mathcal{L}(\phi) = \E_{
        (\aobs, \Act, \Obs) \sim \data
    }
    \left[ \| R_{\phi}(\aobs, \Act) - R(\aobs, \Act, \Obs) \|^2 \right].
\end{align}
\end{definition}

We then alternate between training the reward surrogate and the policy, as described in \Cref{alg:rwfm}. Note that the demonstration action chunks are only used in the reward surrogate loss---not in the \grpo loss. This means that as long as the actions being generated by $\pi_{\theta}$ are correctly evaluated by the reward surrogate, optimizing the \grpo loss will generally improve the policy. Eventually, the policy may produce out-of-distribution action trajectories which are overvalued by the reward surrogate, resulting in poor validation performance. This triggers a round of data collection and produces reward feedback for these learned policy actions. Our algorithm is summarized in \Cref{alg:grpo}.
\section{Experiments}
\begin{figure}
    \centering
    \includegraphics[width=\textwidth]{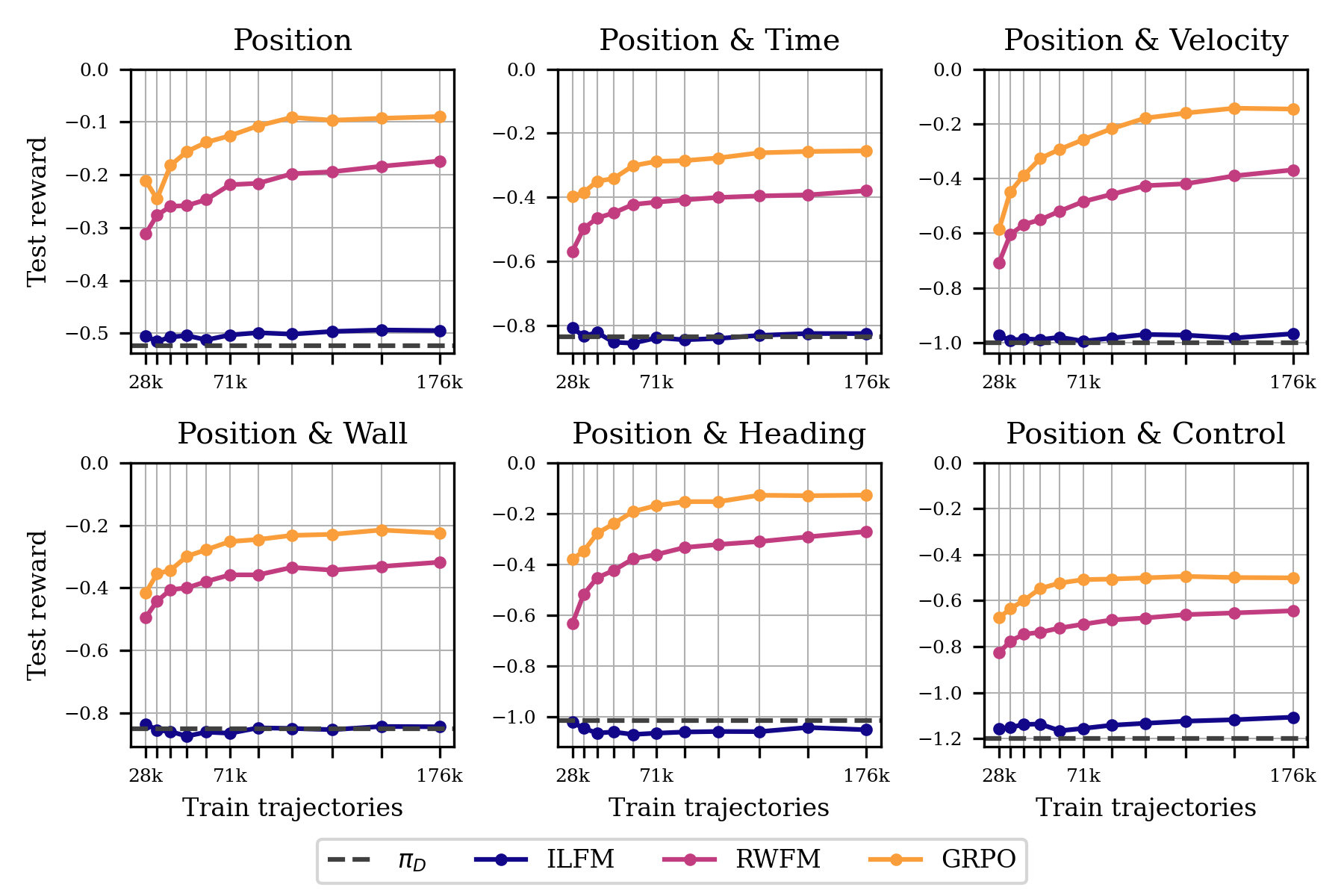}
    \caption{
        Performance comparison of \ilfm, \rwfm, and \grpo policies across different reward functions, with the constant demonstration policy $\pi_D$ performance shown as a gray dotted horizontal line.
        The $x$-axis represents the number of training trajectories, and the $y$-axis shows the average test-time reward.
    }
    \label{fig:main}
\end{figure}

\Cref{sec:experiment_setup} introduces our environment, the suboptimal demonstration policy, and a suite of reward functions. \Cref{sec:results} details our experiment results for the \rwfm and \grpo policies.

\subsection{Environment, demonstrator, and tasks} \label{sec:experiment_setup}
Our environment is an illustrative planar unicycle plant, with a state space consisting of a position $\bp \in \R^2$, a heading vector $\bh \in \R^2$, and a longitudinal velocity $v \in \R$. The control input consists of angular velocity and longitudinal acceleration. Positions are clamped to the $[-1, 1]^2$ meter square after each environment step, and wall collisions immediately reduce longitudinal velocity to zero. Angular velocity is clamped to $[-3, 3]$ rad/s, longitudinal velocity is clamped to $[0, 1]$ m/s, and acceleration is clamped to $[-1, 1]$ m/s$^2$. The environment initial state is uniformly sampled from the valid ranges. We simulate with a timestep of $\Delta t = 0.1$ seconds. The unicycle state is fully observable, and the augmented observation $\aobs$ includes a goal position $\bg \in \R^2$ randomly sampled from $[-1, 1]^2$.

We design a suboptimal demonstrator policy $\pi_D(\Act \mid \obs)$ whose performance we aim to later surpass with reinforcement learning. Our policy is parameterized by a constant goal velocity sampled from $\Unif(0.2, 1.0)$, a heading proportional gain sampled from $\Unif(0.3, 3.0)$, and a velocity proportional gain sampled from $\Unif(0.3, 3.0)$. At each iteration, the demonstrator policy computes the angular heading error to the target position and difference between the current velocity and the goal velocity. These errors are scaled by the proportional gains to produce the angular velocity and acceleration commands. We execute the demonstrator policy for random horizon lengths between $1$ and $64$.

We consider six distinct reward functions $R(\aobs, \Act, \Obs)$, chosen to reflect a diverse range of possible real-world learning objectives; their definitions are provided in \Cref{app:reward_functions}. Our architecture and hyperparameters are provided in \Cref{app:architectures}.

\subsection{Results} \label{sec:results}
\begin{figure}
    \centering
    \includegraphics[width=\textwidth]{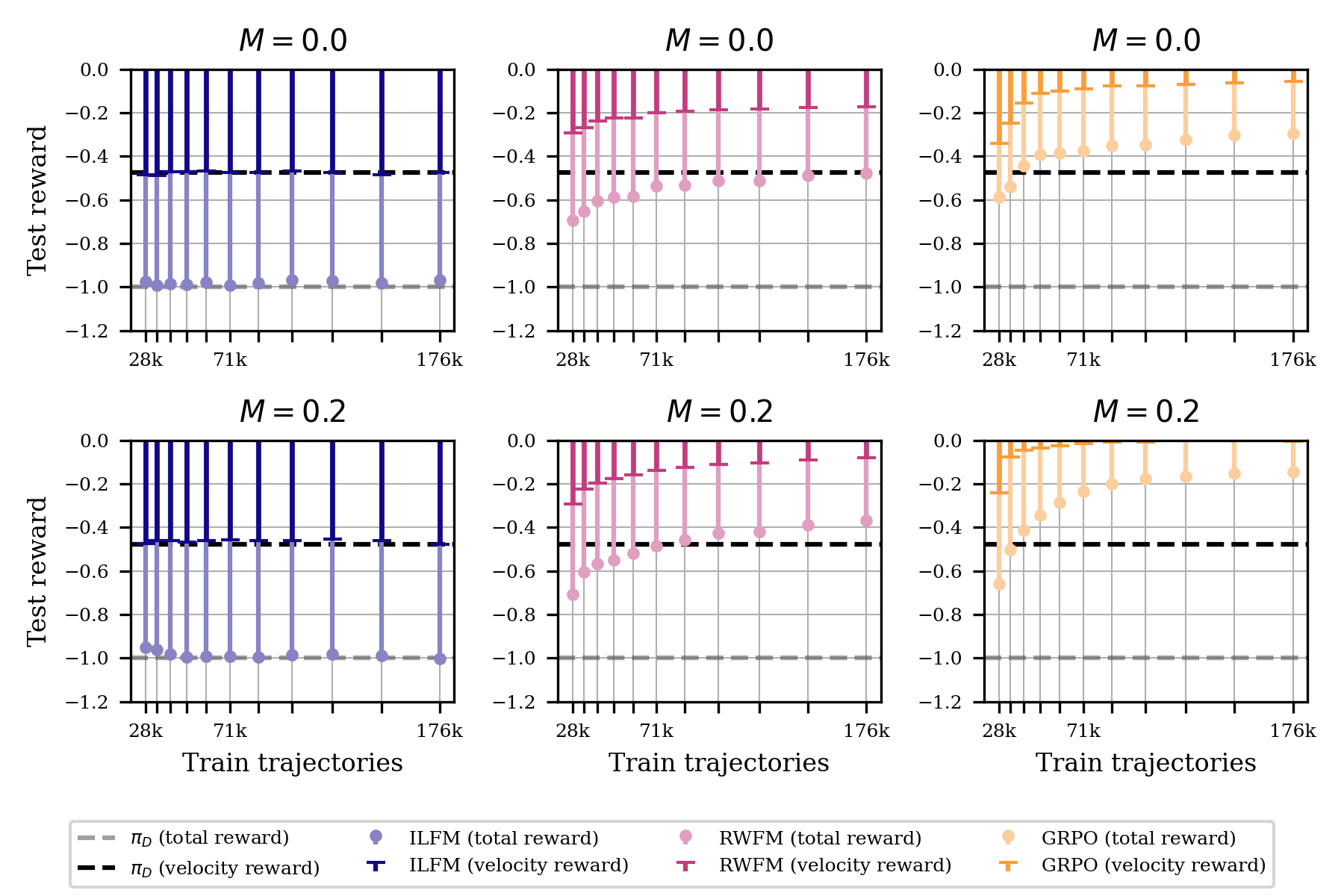}
    \caption{
        Performance comparison of exploration magnitudes $M=0.0$ and $M=0.2$ for the Position \& Velocity reward function.
        The constant demonstration policy $\pi_D$ performance is shown using dotted horizontal lines.
        The $x$-axis represents the number of training trajectories, and the $y$-axis shows the average test-time reward.
        The reward is decomposed as a stacked chart into the velocity reward component (dark shading) and the position plus velocity reward (light shading).
    }
    \label{fig:velocity_optimize}
\end{figure}

\Cref{fig:main} displays performance curves for the reward functions described in \Cref{sec:experiment_setup}.
For the \rwfm policy we sweep over the reward scaling factor $\alpha \in \{0, 5, 10, 20, 40\}$ and action exploration magnitude $M \in \{0.0, 0.05, 0.1, 0.2\}$.
For the \grpo policy we fix $\alpha = 2.0$ and sweep over the same set of action exploration magnitudes.
\Cref{fig:main} uses the best performing hyperparameters for each reward function on a validation set.

\Cref{fig:main} shows that the \ilfm policy performance matches that of the demonstration policy $\pi_D$ as expected, and the \rwfm and \grpo policy performances increase with the number of training trajectories.
The \grpo policy also consistently outperforms the \rwfm policy for all reward functions.
Both policies improve substantially over the suboptimal demonstration policy $\pi_D$.

\Cref{fig:rwfm_alpha_sweep} shows how the \rwfm policy performance changes with different reward scaling factors $\alpha$. Note that $\alpha=0$ corresponds to the vanilla \ilfm policy.
All positive $\alpha$ values significantly improve performance over the \ilfm policy, with best or competitive performance generally achieved for $\alpha \in \{10, 20\}$. Some loss functions show a regression for larger values of $\alpha$. We suspect this may be due to the \rwfm policy being insufficiently sensitive to small changes in the conditioning observation $\aobs$.
To see this, consider the limit of $\alpha \to \infty$, and assume that the highest-reward sample from $\data$ has a reward of $0$ and all others are strictly negative. In this case, the weight function $w$ in \Cref{def:rwfm_loss} becomes a Dirac delta function at the highest-reward sample, and the resulting flow-matching model is likely to reproduce only one action chunk regardless of the conditioning observation $\aobs$.
Thus for $\rwfm$ policies, $\alpha$ induces a trade-off between greedily reproducing high-reward action chunks (potentially compromising on the relevance of $\aobs$) and faithfully reproducing trajectories from $\data$ best suited for the conditioning observation $\aobs$ (without any discernment of which trajectories achieve high rewards).
This trade-off is not present for the \grpo policy, as $\alpha$ weights between samples which are all generated from a particular $\aobs$.

\Cref{fig:rwfm_explore_amplitude_sweep} and \Cref{fig:grpo_explore_amplitude_sweep} in \Cref{app:additional_results} show how the \rwfm and \grpo policy performance changes with different action exploration magnitudes $M$.
Recall that the \rwfm action exploration occurs during trajectory collection, while the \grpo action exploration occurs when sampling trajectories to be scored by the reward surrogate.
For the \grpo policy, we find that a high exploration magnitude $M=0.2$ consistently maintains or improves performance in the large-sample regime.
The \rwfm policy experiences more variable benefit from exploration, with most improvement occurring for the Position \& Velocity reward function.
We suspect that this is due to the fact that achieving zero final velocity requires a ``braking'' behavior beyond the support of $\pi_D$, which only regulates to a fixed target velocity.
We investigate this further in \Cref{fig:velocity_optimize}.
For a low exploration magnitude $M=0.0$, there remains a negative final velocity reward for both the \rwfm and \grpo policies.
This is substantially improved for an exploration magnitude of $M=0.2$, with the \grpo policy in particular achieving a near-zero final velocity reward, indicating that it has successfully learned a novel braking behavior.

\begin{figure}
    \centering
    \includegraphics[width=\textwidth]{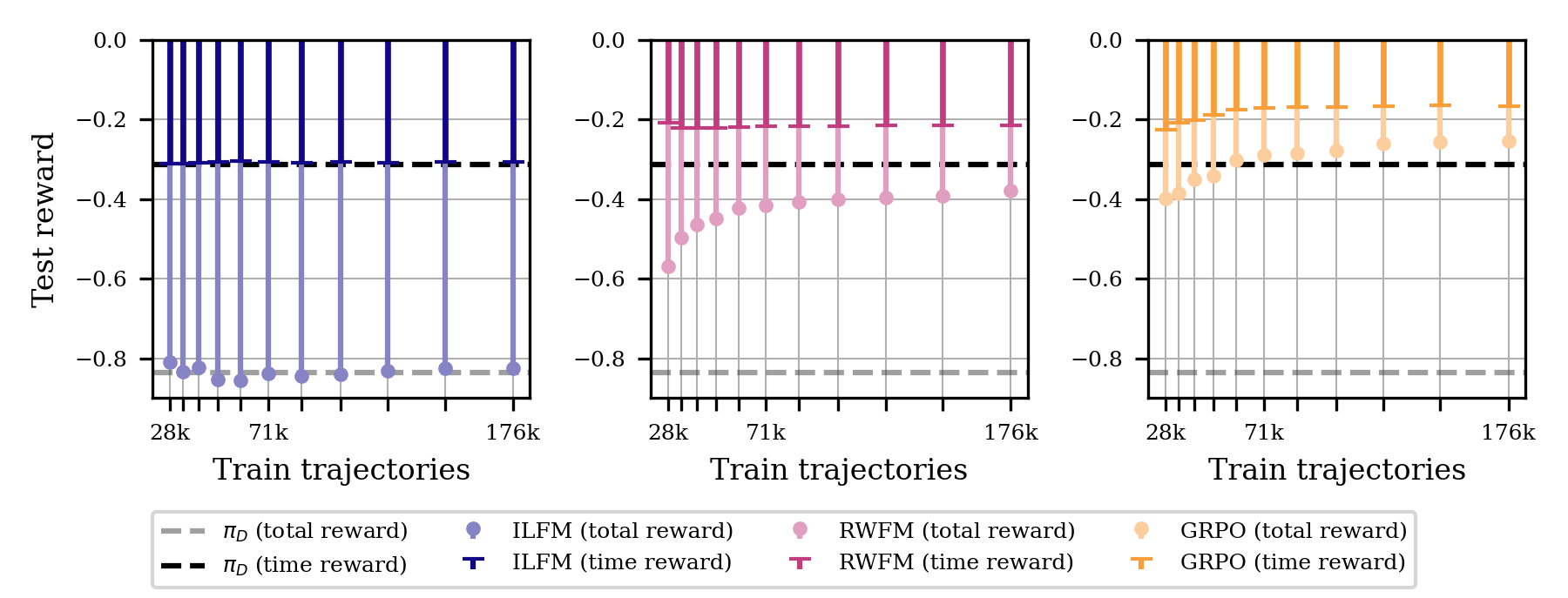}
    \caption{
        Performance comparison of \ilfm, \rwfm, and \grpo policies for the Position \& Time reward function, with the constant demonstration policy $\pi_D$ performance shown as dotted horizontal lines.
        The $x$-axis represents the number of training trajectories, and the $y$-axis shows the average test-time reward.
        The reward is decomposed as a stacked chart into the time reward component (dark shading) and the position plus time reward (light shading).
    }
    \label{fig:time_optimize}
\end{figure}

\Cref{fig:time_optimize} specifically examines the Position \& Time reward function. We show that both the \rwfm and \grpo policies are able to significantly reduce the time reward component beyond the performance of the \ilfm and demonstration policies. This indicates that our scheme for variable-time planning detailed in \Cref{sec:method} enables training agents which act faster than the demonstration policy.

\section{Conclusion}
This work examines two approaches for training flow matching policies via reinforcement learning: a simple reward-weighted flow matching scheme and a group relative policy optimization approach with a learned reward surrogate.
We further introduce the complementary concepts of \emph{variation suboptimality} and \emph{support suboptimality}, which are common in demonstration datasets.
To overcome support suboptimality, we introduce schemes for exploring in the space of action trajectories and allowing flow-matching policies to plan with variable-time horizons.
We evaluate these policies in an illustrative suite of simulated unicycle dynamics tasks and show that our learned policies substantially improve upon the performance of the suboptimal demonstration policy.

\bibliography{references}
\bibliographystyle{plainnat}

\newpage
\appendix

\section{Experimental details}

\subsection{Architectures and hyperparameters} \label{app:architectures}

\inlinesubsection{Data collection}
We start by collecting $30,000$ demonstration trajectories from $\pi_D$. These are held fixed for all remaining experiments, and are split $95\%$ for training, $1\%$ for validation, and $4\%$ for testing.
We define a performance plateau as being unable to exceed the previous best validation performance by $0.01$ for $500$ consecutive epochs for \rwfm and $50$ consecutive epochs for \grpo. We perform $C=10$ collection iterations with a sampling fraction of $\gamma=0.2$.

\inlinesubsection{Flow matching model}
Our flow-matching velocity field $\bv_{\theta}$ is a standard convolutional U-Net, repurposed from \citet{lipman2024flowmatchingguidecode}. 
We train with a learning rate of $5 \cdot 10^{-3}$ using the AdamW optimizer with $\beta_1=0.9$ and $\beta_2=0.95$.
The batch size is $512$ for the \ilfm and \rwfm methods. For the \grpo method, we drop the batch size to $128$.

Our \grpo implementation uses a group size $G=10$ and $\alpha=2.0$.

To generate samples, we use Euler integration with $4$ steps.

\inlinesubsection{Reward surrogate}
Our reward surrogate $R_{\phi}$ is derived from the TimesNet architecture for time-series classification \citep{wu2023timesnet}. The input sequence consists of concatenating $\aobs \in \R^{d_{\aobs}}$ to each action $\act[i] \in \R^{d_a}$ in $\Act$, resulting in a tensor of shape $\R^{(d_a + d_{\aobs}) \times H}$. We then modify the output size to a single logit and use that as our regression output for $R_{\phi}$. We use $E_{rs}=3$ reward surrogate training sub-epochs with a learning rate of $10^{-4}$ using the AdamW optimizer with $\beta_1=0.9$ and $\beta_2=0.95$. The batch size is $512$.

\subsection{Reward functions} \label{app:reward_functions}
Let $\obs_{H+1}$ denote the last observation of the rollout $\Obs$, with $H$ denoting the horizon of the actions $\Act$ after interpolation of $\hat{\Act}$ to the original timestep of $\Delta t = 0.1$. Our reward functions are then defined as follows:

\textbf{Final position (Position).} This reward function penalizes the distance between the final position and the goal: $R(\aobs, \Act, \Obs) = -\|\bp_{H+1} - \bg\|_2$.

\textbf{Final position and total time (Position \& Time).} This adds a total time elapsed penalty to the final position reward: $R(\aobs, \Act, \Obs) = -\|\bp_{H+1} - \bg\|_2 - 0.1 H \Delta t$, where $0.1$ is a scaling constant.

\textbf{Final position and final velocity (Position \& Velocity).} This adds a final velocity penalty to the final position reward, encouraging the agent to come to a stop near the goal: $R(\aobs, \Act, \Obs) = \|\bp_{H+1} - \bg\|_2 - v_{H+1}$.

\textbf{Final position and wall collisions (Position \& Wall).} This adds a wall collision penalty to the final position reward: $R(\aobs, \Act, \Obs) = -\|\bp_{H+1} - \bg\|_2 - \text{Wall}(\Obs)$, where $\text{Wall}(\Obs)$ is $1$ if the agent collides with a wall at some point during the rollout and zero otherwise.

\textbf{Final position and final heading (Position \& Heading).} This adds a vertical heading penalty to the final position reward, encouraging the agent to point upwards at the end of the trajectory: $R(\aobs, \Act, \Obs) = -\|\bp_{H+1} - \bg\|_2 - \frac{1}{2} (1 - \bh_y)$. We let $\bh_y$ denote the $y$-axis component of $\bh$ and scale and shift the additional penalty to lie within $[-1, 0]$.

\textbf{Final position and control regularization (Position \& Control).} This adds a maximum control penalty to the final position reward: $R(\aobs, \Act, \Obs) = -\|\bp_{H+1} - \bg\|_2 - \frac{1}{d_a} \big(1 - \sum_{i=1}^{d_a} A_i^{absmax}\big)$, where $A_i^{absmax}$ denotes the maximum absolute value of the $i$-th control channel in $\Act$ across the planning horizon, after normalizing by the actuation bounds to shift values into the $[-1, 1]$ range.

\section{Additional results} \label{app:additional_results}
This section contains additional experimental results sweeping the reward scaling factor $\alpha$ and action exploration magnitude $M$.

\begin{figure}[H]
    \centering
    \includegraphics[width=\textwidth]{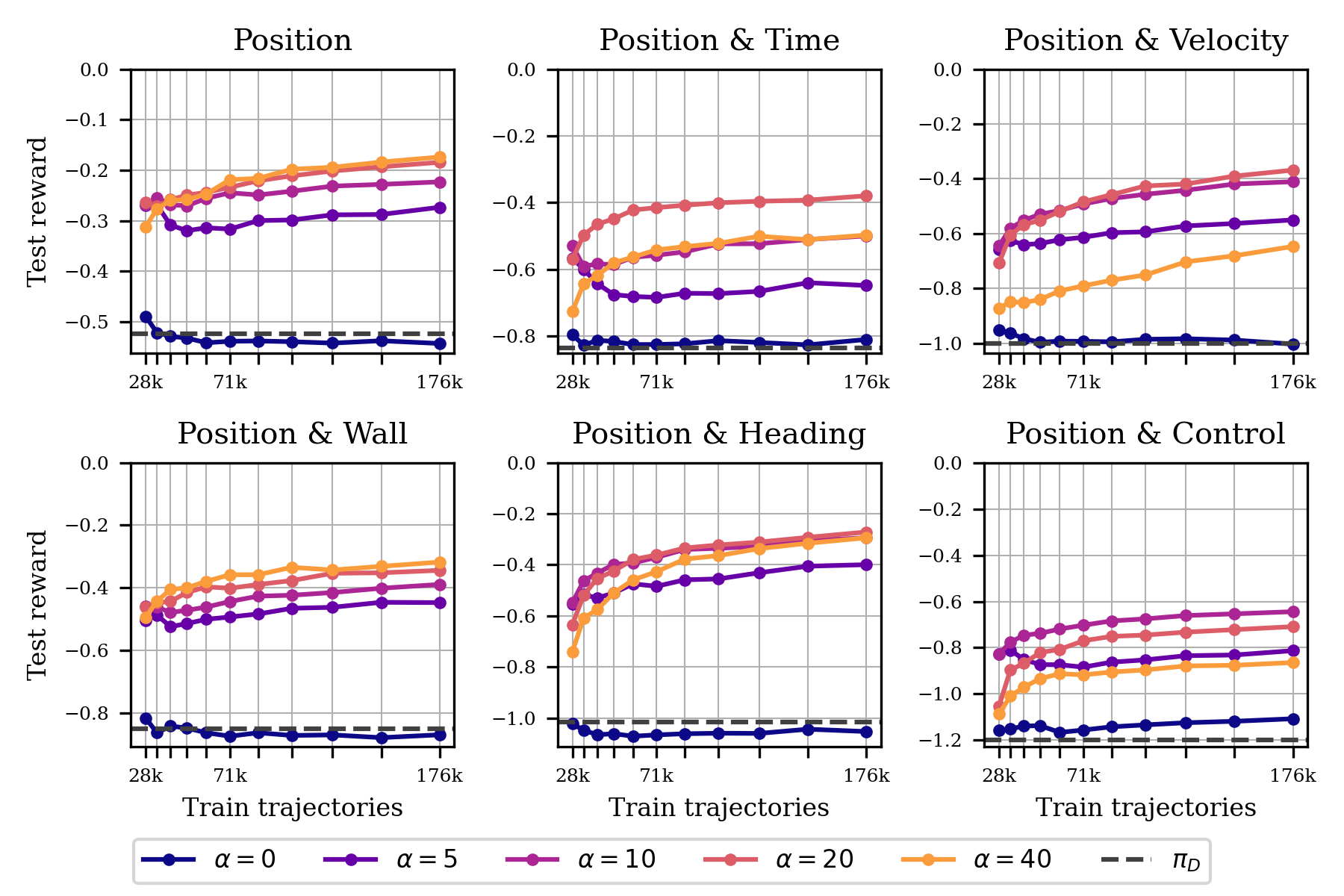}
    \caption{
        \rwfm policy performance comparison with different reward scaling factors $\alpha$.
    }
    \label{fig:rwfm_alpha_sweep}
\end{figure}

\begin{figure}[H]
    \centering
    \includegraphics[width=\textwidth]{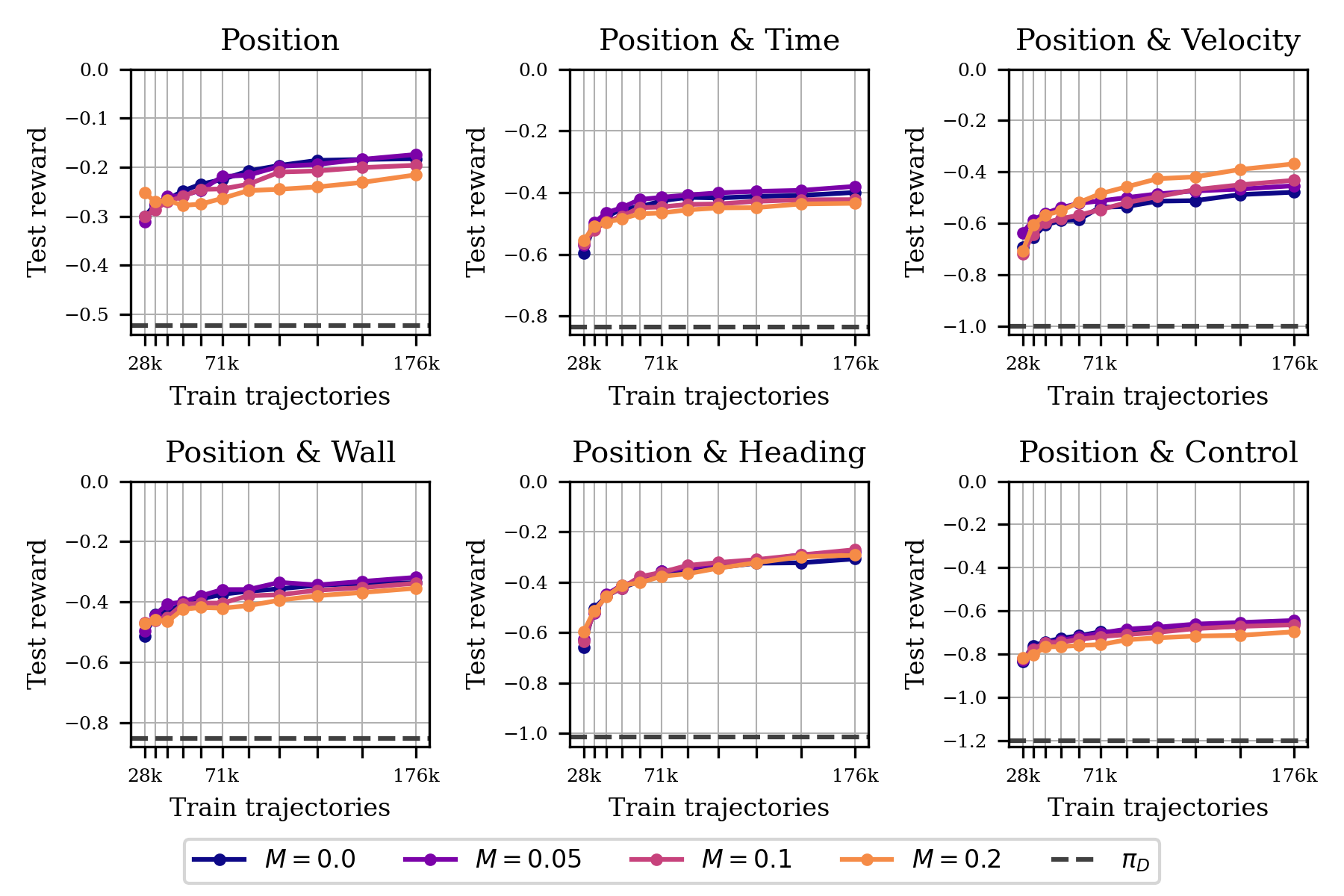}
    \caption{
        \rwfm policy performance comparison with different action exploration magnitudes $M$.
    }
    \label{fig:rwfm_explore_amplitude_sweep}
\end{figure}

\begin{figure}[H]
    \centering
    \includegraphics[width=\textwidth]{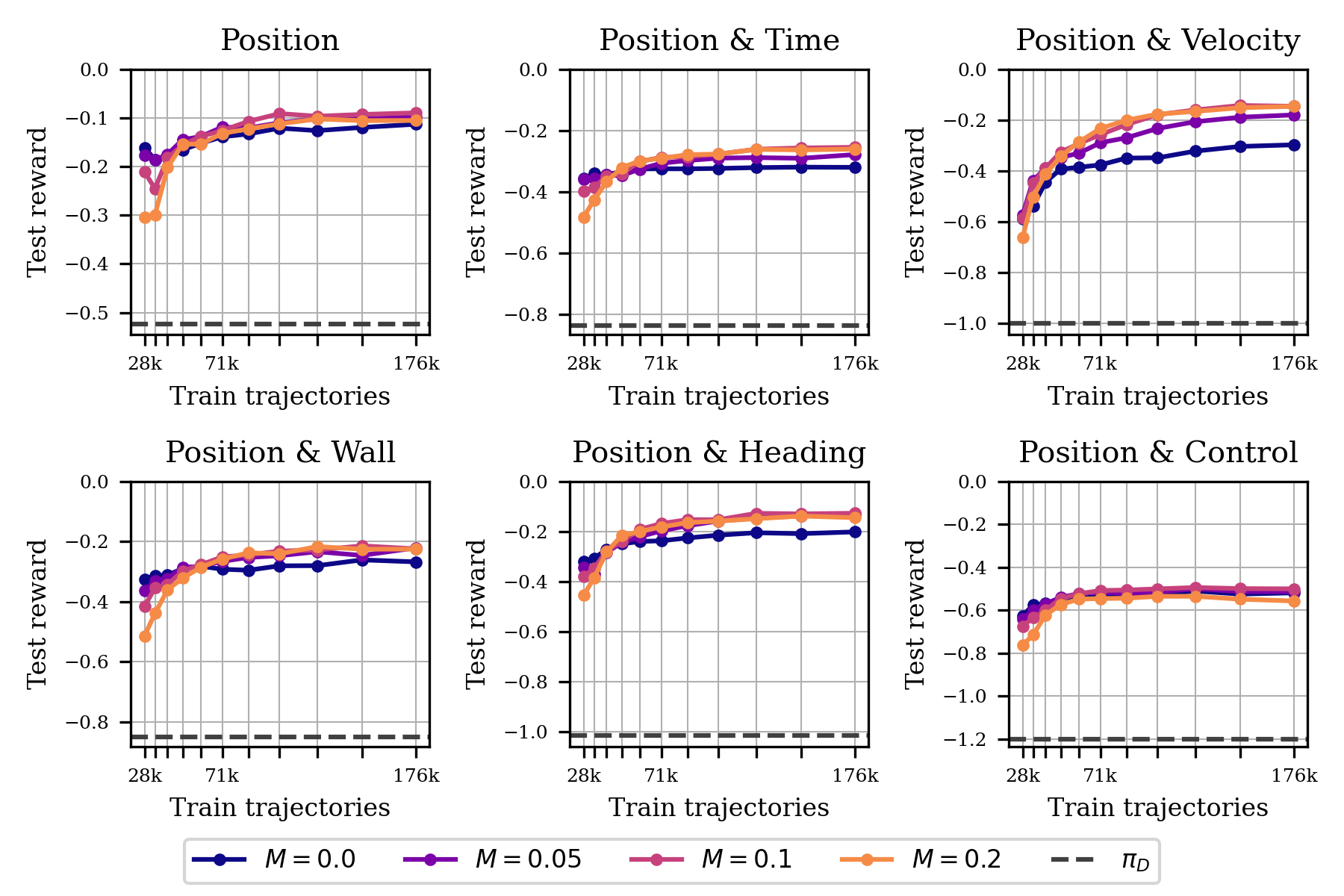}
    \caption{
        \grpo policy performance comparison with different action exploration magnitudes $M$.
    }
    \label{fig:grpo_explore_amplitude_sweep}
\end{figure}

\end{document}